\documentclass{article}
\usepackage{spconf,amsmath,graphicx,booktabs,array,tabularx}
\newcolumntype{L}[1]{>{\raggedright\arraybackslash}m{#1}}
\newcolumntype{C}[1]{>{\centering\arraybackslash}p{#1}}

\let\OLDthebibliography\thebibliography
\renewcommand\thebibliography[1]{
  \OLDthebibliography{#1}
  \setlength{\parskip}{0pt}
  \setlength{\itemsep}{0pt plus 0.3ex}
}

\usepackage{cite}
\usepackage{algorithmic}
\usepackage{float}
\usepackage[linesnumbered,ruled,vlined]{algorithm2e}

\title{PRESERVE PRE-TRAINED KNOWLEDGE: TRANSFER LEARNING WITH SELF-DISTILLATION FOR ACTION RECOGNITION}
\name{Yang Zhou, Zhanhao He, Keyu Lu, Guanhong Wang, Gaoang Wang$^{\ast}$}
\address{Zhejiang University-University of Illinois at Urbana-Champaign Institute, Zhejiang University, China\\
\{yangz.19, zhanhao.19, keyu.19, gaoangwang\}@intl.zju.edu.cn, guanhongwang@zju.edu.cn}

\begin{document}
%

\maketitle
\renewcommand{\thefootnote}{\fnsymbol{footnote}}

\footnotetext[1]{Corresponding author.}
\begin{abstract}

Video-based action recognition is one of the most popular topics in computer vision. 
With recent advances of self-supervised video representation learning approaches, action recognition usually follows a two-stage training framework, \textit{i.e.}, self-supervised pre-training on large-scale unlabeled sets and transfer learning on a downstream labeled set. 
However, catastrophic forgetting of the pre-trained knowledge becomes the main issue in the downstream transfer learning of action recognition, resulting in a sub-optimal solution.
In this paper, to alleviate the above issue, we propose a novel transfer learning approach that combines self-distillation in fine-tuning to preserve knowledge from the pre-trained model learned from the large-scale dataset. Specifically, we fix the encoder from the last epoch as the teacher model to guide the training of the encoder from the current epoch in the transfer learning. 
With such a simple yet effective learning strategy, we outperform state-of-the-art methods on widely used UCF101 and HMDB51 datasets in action recognition task.
\end{abstract}
\section{Introduction}
\label{sec:intro}

Action recognition is a fundamental topic of computer vision. It has gained significant attention due to its wide range of applications in video surveillance, video recommendation, and human-computer interaction \cite{huang2018makes}. 
Early works \cite{simonyan2014twostream,tran2015learning} rely on a large amount of labeled data for learning spatial-temporal representations, requiring huge annotation costs. To alleviate the dependence on the labeled data, pre-training strategies for self-supervised video representation learning have attracted extensive attention. These approaches, such as IIC\cite{IIC} and CVRL\cite{CVRL}, require only a large-scale set with few or no annotations. 


After self-supervised pre-training, usually, transfer learning is employed for downstream tasks on the labeled set. However, 
catastrophic forgetting of the pre-trained knowledge would become the main issue in the transfer learning stage. When adapting the model to the labeled set, the knowledge of large-scale unlabeled data will be gradually forgotten. 

To address the above issue, we propose a novel transfer learning method that combines self-distillation in the fine-tuning stage. Specifically, at the beginning of each training epoch, we make a copy of the base encoder network from the last epoch as the teach model and freeze the model parameters. Except for the conventional cross-entropy loss used for classification in training, we constrain the difference between representations from the teacher model and the base encoder with the Euclidean distance as the regularization term. As a result, the previous knowledge of the model can be largely preserved. With such a simple yet effective transfer learning approach, we achieve 82.3$\%$ and 47.3$\%$ top-1 accuracy on UCF101\cite{soomro2012ucf101} and HMDB51\cite{Kuehne11} datasets, respectively, outperforming state-of-the-art methods.

In summary, the main contributions of this paper are as follows: 
\begin{itemize}
\item We propose a novel self-distillation guided transfer knowledge approach for action recognition that can alleviate the catastrophic forgetting phenomenon and preserve the pre-trained knowledge learned from large-scale datasets. 
\item Our method can be fitted for different architectures and models, and easily applied to any existing transfer learning frameworks. 
\item We achieve state-of-the-art performance on commonly used action recognition datasets to demonstrate the effectiveness of our proposed method.
\end{itemize}

\begin{figure*}[t]
    \centering
    \includegraphics[width=0.98\textwidth]{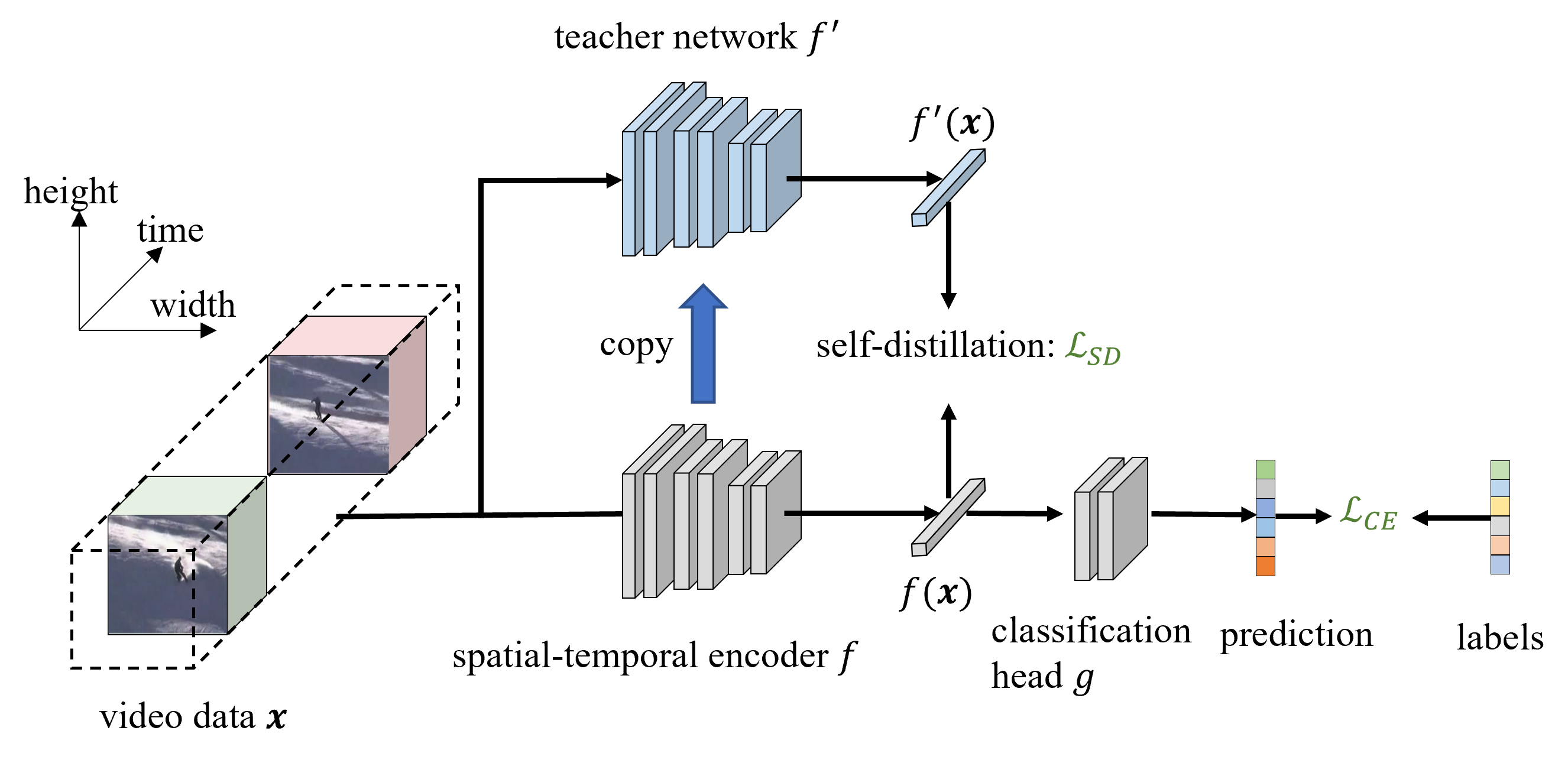}
    \caption{Overview of our proposed self-distillation guided transfer learning framework. At the beginning of each epoch, we make a copy of the spatial-temporal encoder from the last epoch as the teacher network $f'(\cdot)$. The Euclidean distance between representations of both encoders is treated as a regularization in transfer learning. Besides that, the conventional cross-entropy loss is employed on the head network $g(\cdot)$ for action recognition.}
    \label{fig:framework}
\end{figure*}

\section{Related Work}
\label{sec:format}

\subsection{Encoder Models for Action Recognition}
Encoder models play an essential role in representation learning in video-based action recognition. To learn spatial-temporal information of videos, different types of encoding architecture are widely adopted, including two-stream networks, CNN-LSTM networks, 3D CNN, and vision transformers.
For example, the two-stream architecture, which was first proposed in \cite{simonyan2014twostream}, incorporates spatial and temporal streams based on CNN. 
\cite{donahue2015long} presents LRCN, which combines CNN and LSTM to learn the spatial features and temporal features more thoroughly. 
\cite{tran2015learning} proposes 3D CNN to learn spatial-temporal features directly. It performs better compared with 2D CNN. 
Moreover, transformer-based encoder has been applied to action recognition and achieves brilliant results. For example, \cite{chen2022mm} handles the large number of spatial-temporal tokens extracted from multiple modalities and promotes learning the inter-modal interactions. 
Additionally, in recent years, light-weighted models with less computational cost have been explored in action recognition. Specifically, \cite{s3d} replaces computational expensive 3D convolutions with lower-cost 2D convolutions. 
\cite{slowfast} proposes SlowFast model, which applies a Slow pathway to capture spatial semantics and applies a Fast pathway to capture temporal information, like a two-stream model. \cite{Tran2019} suggests factorizing 3D convolutions by separating channel interactions and spatial-temporal interactions. 
\cite{Feichtenhofer_2020_CVPR} expands a tiny 2D image classification architecture. 
The current study shows that it is still a challenge to learn good representations with long periods of video actions. Without guidance on action labels, the encoders of action recognition learn background features rather than foreground actions. They analyze the background to assist action recognition, which deviates from the purpose of the representation learning task.

\subsection{Self-Supervised Learning for Action Recognition Pre-Training}
With a large-scale unlabeled dataset, self-supervised video representation learning has gained more attention from the community. For example,  
\cite{CVRL} presents a self-supervised video representation learning method with contrastive loss, where two augmented clips from the same short video are pulled together, and clips from different videos are pushed away. \cite{chen2021rspnet} presents a new way to detect the playback speed and exploits the relative speed between two video clips as labels. \cite{chen2021rspnet} also proposes to enforce the model to capture the appearance difference between two video clips. \cite{Zhang_2022_WACV} decomposes the representation learning objective into two sub-tasks, respectively, emphasizing spatial and temporal features. \cite{Wang_2019_CVPR} proposes to track cycle-consistency as free supervisory signal for learning visual representations. 
\cite{han2020memory} proposes a new architecture and learning framework named Memory-augmented Dense Predictive Coding for self-supervised video representation learning task. It trains the network with a predictive attention mechanism over the set of compressed memories and represents the future states with the combination of condensed representations in memory.
Current pre-training-based action recognition methods usually adopt simple transfer learning on the downstream labeled dataset. Designing more effective transfer learning approaches to address the catastrophic forgetting phenomenon of pre-trained knowledge is still under-investigated.

\subsection{Self-Distillation}
Knowledge distillation was firstly proposed in \cite{hinton2015distilling} to transfer the knowledge from
the cumbersome model to a small model.
Self-distillation can be treated as a regularization strategy when the teacher model and the student model share the same architecture. For example, \cite{zhang2019your} first uses self-distillation, 
which transfers the knowledge in the deeper portion of the networks to the shallow sections.
\cite{bhat2021distill} improves representation quality of the smaller models by single-stage online knowledge distillation.
\cite{kim2021self} uses self-distillation between different epochs with soft targets. However, combining self-distillation as the regularization for transfer learning is not well-exploited.

\section{Method}
\label{sec:pagestyle}
In this section, we demonstrate our proposed self-distillation guided transfer learning framework for action recognition in detail. We first show the conventional transfer learning framework for downstream tasks with pre-trained encoder network. Then, we illustrate how to incorporate the self-distillation to adapt the network on the downstream dataset and memorize the pre-trained knowledge simultaneously.

\subsection{Transfer Learning with Pre-Trained Model}
Denote $f_{\boldsymbol{\theta}}(\cdot)$ as the base encoder network with parameters $\boldsymbol{\theta}$ that are pre-trained on the large-scale unlabeled dataset via self-supervised learning \cite{chen2021rspnet}. 
To achieve the goal of action recognition as the downstream task, we stack the base encoder with an extra classification head network $g_{\boldsymbol{\phi}}(\cdot)$, where $\boldsymbol{\phi}$ represents the parameters of the head network. Then we apply transfer learning to the entire network with the following cross-entropy loss,
\begin{equation}
    \mathcal{L}_{CE}(\mathcal{X}|\boldsymbol{\phi},\boldsymbol{\theta}) = -\sum_{i=1}^{N}\log(g^{y_i}_{\boldsymbol{\phi}}(f_{\boldsymbol{\theta}}(\boldsymbol{x}_i))),
\label{eq:ce_loss}
\end{equation}
where $\mathcal{X}$ represents the training data, $(\boldsymbol{x}_i, y_i)$ is the individual data-label pair,  $g^{y_i}_{\boldsymbol{\phi}}(f_{\boldsymbol{\theta}}(\boldsymbol{x}_i))$ represents the prediction of the ground truth class $y_i$ after softmax operation, and $N$ is the number of training samples. We can obtain the learned parameters $\hat{\boldsymbol{\phi}}$ and $\hat{\boldsymbol{\theta}}$ by minimizing the above loss. 
 
\subsection{Self-Distillation Guided Transfer Learning}
Eq.~(\ref{eq:ce_loss}) is the conventional transfer learning for the downstream task. However, it has the issue of catastrophic forgetting of the pre-trained knowledge. To address the above issue, we propose a self-distillation guided transfer learning, aiming to adapt the network on the downstream dataset and memorize the pre-trained knowledge simultaneously. Specifically, at the beginning of each epoch $t$, we make a copy of the encoder network from the last epoch $t-1$ and treat the encoder $f'_{\boldsymbol{\theta}^{t-1}}(\cdot)$ as the teacher model. We fix the parameters of the teacher model at the current epoch in training. Then the self-distillation is defined between the encoder model and the teacher model as follows,  
\begin{equation}
    \mathcal{L}_{SD}(\mathcal{X}|\boldsymbol{\theta}^{t}) = \sum_{i=1}^{N}||f_{\boldsymbol{\theta}^{t}}(\boldsymbol{x}_i)-f'_{\boldsymbol{\theta}^{t-1}}(\boldsymbol{x}_i)||^2. 
\label{eq:sd_loss}
\end{equation}
The base encoder model plays a role as the student model in the self-distillation. The self-distillation is treated as a regularization term in transfer learning and used for memorizing the knowledge from the last epoch. At the first epoch, \textit{i.e.}, $t=1$, we use the pre-trained base encoder as the teacher model $f'_{\boldsymbol{\theta}^{0}}(\cdot)$. 

Combined with cross-entropy loss in Eq.~(\ref{eq:ce_loss}) and self-distillation loss in Eq.~(\ref{eq:sd_loss}), the total loss $L$ for the $t$-th epoch can be written as follows,
\begin{equation}
    \mathcal{L}(\mathcal{X}|\boldsymbol{\phi}^{t},\boldsymbol{\theta}^{t}) = \mathcal{L}_{CE}(\mathcal{X}|\boldsymbol{\phi}^{t},\boldsymbol{\theta}^{t})+\lambda \mathcal{L}_{SD}(\mathcal{X}|\boldsymbol{\theta}^{t}),
\label{eq:loss}
\end{equation}
where $\lambda$ is the weight for self-distillation loss. The training details are summarized in Algorithm~\ref{alg:train}.

\begin{algorithm}
  \textbf{Input}: Labeled set $\mathcal{X}$ and pre-trained model $f_{\boldsymbol{\theta}}$.\\
  
  \textbf{Initialization}: Initialize head classifier $g_{\boldsymbol{\phi}}$. \\
  
  Set $\boldsymbol{\theta}^0\leftarrow \boldsymbol{\theta}$. \\
  \For{$t=1,2,...,t_{max}$} 
  {
    $f'_{\boldsymbol{\theta}^{t-1}} \leftarrow f_{\boldsymbol{\theta}^{t-1}}$ and freeze  $f'_{\boldsymbol{\theta}^{t-1}}$.\\
    \For{Batch data $\mathcal{B}$ in $\mathcal{X}$}
    {   
        $\boldsymbol{\theta}^{t} \leftarrow \boldsymbol{\theta}^{t-1} - \eta \boldsymbol{\bigtriangledown}_{\boldsymbol{\theta}^{t-1}} \mathcal{L}(\mathcal{B}|\boldsymbol{\phi}^{t},\boldsymbol{\theta}^{t})$.\\
        
        $\boldsymbol{\phi}^{t} \leftarrow \boldsymbol{\phi}^{t-1} - \eta \boldsymbol{\bigtriangledown}_{\boldsymbol{\phi}^{t-1}} \mathcal{L}(\mathcal{B}|\boldsymbol{\phi}^{t},\boldsymbol{\theta}^{t})$.\\
    }
  }
  
  Set $\hat{\boldsymbol{\theta}} \leftarrow {\boldsymbol{\theta}^{t_{max}}}$, $\hat{\boldsymbol{\phi}} \leftarrow {\boldsymbol{\phi}^{t_{max}}}$. \\
  
  \textbf{Output}: $\hat{\boldsymbol{\theta}}, \hat{\boldsymbol{\phi}}$.\\
 
\caption{Training}
\label{alg:train}
\end{algorithm}

\section{Experiments and Results}
\label{sec:typestyle}
\subsection{Datasets}
We adopt top-1 and top-5 accuracy (denoted as Acc@1 and Acc@5, respectively) as evaluation metrics. To verify the effectiveness of our proposed method, we use UCF101\cite{soomro2012ucf101} and HMDB51\cite{Kuehne11} datasets in the evaluation. The statistics of the two datasets are summarized as follows.

\textbf{UCF101} \cite{soomro2012ucf101} is a widely used dataset which contains 13,320 clips with 101 action categories. Following previous work \cite{xu2019self}, we use the training split-1 as the finetuning dataset and testing split-1 for evaluation.

\textbf{HMDB51} \cite{Kuehne11} is a dataset which consists of 6,849 clips with 51 human action classes from YouTube. It is divided into three training/testing splits. Following prior work \cite{xu2019self}, we use training/testing split-1 in the experiment.


\subsection{Implementation Details}
We adopt the pre-trained model based on RSPNet \cite{chen2021rspnet}. The feature generated by spatial-temporal encoder $f_{\boldsymbol{\theta}}(\cdot)$ is of size $24\times512$ and $32\times512$ for C3D\cite{tran2015learning} and R(2+1)D\cite{tran2018closer}, respectively. Classification head $g_{\boldsymbol{\phi}}(\cdot)$ is composed of one fully connected layer. In the transfer learning stage, we use stochastic gradient descent (SGD) optimizer with learning rate 0.005 for C3D model and 0.1 for R(2+1)D model. All the models are fine-tuned for 100 epochs. The setting of the hyper-parameter $\lambda$ is illustrated in the ablation study.

\begin{table}[H]
\begin{center}
\caption{Performance on UCF101 and  HMDB51 compared with RSP. * symbol indicates that we fine-tune 100 epochs on original RSP basis.} 
\label{tab:tabel1}
\begin{tabular}{L{2.1cm}|C{1.05cm}C{1.05cm}|C{1.05cm}C{1.05cm}}
  \toprule
  & \multicolumn{2}{c}{UCF101} & \multicolumn{2}{c}{HMDB51}\\
  Method & Acc@1 & Acc@5 & Acc@1 & Acc@5\\
  \midrule
  \multicolumn{5}{l}{\textbf{C3D Architecture}}\\
  \midrule
  3D ST \cite{kim2019self}
& 60.6 & - & 28.3 & - \\

MAS  \cite{wang2019self} & 61.2 & - & 33.4 & - \\
RSPNet \cite{chen2021rspnet}   & 76.7 & - & 44.6 & - \\
  RSPNet*\cite{chen2021rspnet}   & 77.1 & 93.5 & 43.9   & 76.5  \\
  \textbf{Ours} & \textbf{78.6} & \textbf{94.6} & \textbf{45.8} & \textbf{76.5} \\
  \midrule
  \multicolumn{5}{l}{\textbf{R(2+1)D Architecture}}\\
  \midrule
  Pace \cite{wang2020self}    & 77.1 & - & 36.6 & - \\
  RSPNet \cite{chen2021rspnet}   & 81.1 & - &  44.6  & - \\
  RSPNet* \cite{chen2021rspnet}    & 80.2 & 95.0 &  44.1 & 75.6 \\
  \textbf{Ours} & \textbf{82.3} & \textbf{96.3} & \textbf{47.3} & \textbf{77.8}\\
  \bottomrule
\end{tabular}
\end{center}
\end{table}

\subsection{Comparison with State-of-the-Art Methods}
The comparison with other state-of-the-art methods is shown in Table~\ref{tab:tabel1}, including Acc@1 and Acc@5 results with C3D \cite{tran2015learning} and R(2+1)D \cite{tran2018closer} architectures on UCF101 and HMDB51 datasets, respectively. 
The compared methods include 3D ST \cite{kim2019self}, MAS \cite{wang2019self}, Pace \cite{wang2020self} and RSPNet \cite{chen2021rspnet}. 
Specifically, 3D ST \cite{kim2019self} solves 3D video cubic puzzles by using pretext task in self-supervised video represent learning. 
Instead of designing new self-supervised task, MAS \cite{wang2019self} regresses both motion and appearance statistics along spatial and temporal dimensions to learn visual features. Pace\cite{wang2020self} novelly uses pace prediction to learn spatio-temporal features, and RSPNet \cite{chen2021rspnet} further proposes relative speed perception as a more effective supervision method for video presentation learning.  
For a fair comparison, we re-implement RSPNet with 100-epoch fine-tuning, denoted as RSPNet*, where RSPNet represents originally reported results in \cite{chen2021rspnet} that are based on 30-epoch fine-tuning.  
From the results, our method consistently improves the Acc@1 and Acc@5 performance on two datasets, verifying the effectiveness of our method. Specifically, when using R(2+1) model, we achieve 82.3$\%$ Acc@1 on UCF101 dataset and 47.3$\%$ on HMDB51 dataset, which is significantly higher than other state-of-the-art methods.



\subsection{Ablation Study}
To analyze the influence of the self-distillation loss, we conduct ablation study with different settings of the loss weight $\lambda$. We vary $\lambda$ from 0.1 to 1000 and report Acc@1 and Acc@5 results in Table~\ref{tab:tabel2} with C3D and R(2+1)D architectures on UCF101 and HMDB51 datasets, respectively. 
From the results, even with an extensive range of choices of $\lambda$, our proposed method outperforms the baseline RSPNet in most settings, verifying the effectiveness of the proposed self-distillation guided transfer learning framework.

\begin{table}[H]
\begin{center}
\caption{Top-1 and Top-5 accuracy on UCF101 and HMDB51 datasets. $\lambda$ is varying from 0.1 to 1000.} 
\label{tab:tabel2}
\begin{tabular}{L{2.5cm}|C{1.0cm}C{1.0cm}|C{1.0cm}C{1.0cm}}
  \toprule
  & \multicolumn{2}{c}{UCF101} & \multicolumn{2}{c}{HMDB51}\\
  Method & Acc@1 & Acc@5 & Acc@1 & Acc@5\\
  \midrule
  \multicolumn{5}{l}{\textbf{C3D Architecture}}\\
  \midrule
 RSPNet baseline & 77.1 & 93.5 & 43.9 & 76.5 \\
 Ours ($\lambda$ = 0.1) & 76.7 & 93.7 & 45.8 & 76.5 \\
 Ours ($\lambda$ = 1) & 77.1 & 94.1 & 43.9 & 76.7 \\
 Ours ($\lambda$ = 10) & 77.9 & 94.1 & 44.2 & 76.7  \\
 Ours ($\lambda$ = 100) & 78.6 & 94.6 & 42.9 & 77.4  \\
 Ours ($\lambda$ = 1000) & 77.5 & 94.7 & 41.4 & 75.9  \\
  \midrule
  \multicolumn{5}{l}{\textbf{R(2+1)D Architecture}}\\
  \midrule

 RSPNet baseline & 80.2 & 95.0 & 44.1 & 75.6 \\
 Ours ($\lambda$ = 0.1) & 80.8 & 94.4 & 44.8 & 74.3 \\
 Ours ($\lambda$ = 1) & 80.3 & 94.8 & 43.3 & 75.1 \\
 Ours ($\lambda$ = 10) & 81.4 & 95.4 & 45.6 & 76.1  \\
 Ours ($\lambda$ = 100) & 82.3 & 96.3 & 46.3 & 76.3 \\
 Ours ($\lambda$ = 1000) & 81.5 & 96.1 & 47.3 & 77.8  \\
  \bottomrule
\end{tabular}
\end{center}
\end{table}

\section{Conclusion}
\label{sec:majhead}
In this paper, we propose a novel, simple yet effective transfer learning method with self-distillation to preserve pre-training knowledge for action recognition. The self-distillation is conducted as a regularization term in the fine-tuning to memorize the knowledge learned from the previous model. It alleviates the phenomenon of catastrophic forgetting on the labeled set. Experiments on UCF101 and HMDB51 datasets show that our proposed transfer learning method consistently improves the top-1 and top-5 accuracy for action recognition task.  



\bibliographystyle{IEEEbib}
\bibliography{reference}

\end{document}